\pdfoutput=1

\documentclass[11pt]{article}

\usepackage[preprint]{acl}
\usepackage{tabularx}
\usepackage{stfloats}

\usepackage{times}
\usepackage{caption}
\usepackage{latexsym}
\usepackage{fancyvrb} 
\usepackage{pdflscape}  

\usepackage[T1]{fontenc}

\usepackage{microtype}

\usepackage{inconsolata}

\usepackage{graphicx}
\usepackage{float}
\usepackage{algorithm}

\usepackage[utf8]{inputenc}
\usepackage{amsmath}
\usepackage{algorithm}
\usepackage{algpseudocode}
\usepackage{amssymb}

\usepackage{listings}
\usepackage{xcolor}
\usepackage{enumitem}

\lstdefinelanguage{json}{
    basicstyle=\ttfamily\small,
    showstringspaces=false,
    breaklines=true,
    frame=single,
    backgroundcolor=\color{gray!10},
    literate=
     *{0}{{{\color{blue}0}}}{1}
      {1}{{{\color{blue}1}}}{1}
      {2}{{{\color{blue}2}}}{1}
      {3}{{{\color{blue}3}}}{1}
      {4}{{{\color{blue}4}}}{1}
      {5}{{{\color{blue}5}}}{1}
      {6}{{{\color{blue}6}}}{1}
      {7}{{{\color{blue}7}}}{1}
      {8}{{{\color{blue}8}}}{1}
      {9}{{{\color{blue}9}}}{1}
      {:}{{{\color{red}:}}}{1},
}

\newcommand{\ourmodel}{{\tt RedactOR}}

\title{{RedactOR: An LLM-Powered Framework for Automatic Clinical Data De-Identification}}

\author{
 \textbf{Praphul Singh\thanks{These authors contributed equally to this work.}},
 \textbf{Charlotte Dzialo\footnotemark[1]},
 \textbf{Jangwon Kim},
 \textbf{Sumana Srivatsa},
\\
 \textbf{Irfan Bulu},
 \textbf{Sri Gadde},
 \textbf{Krishnaram Kenthapadi} 
\\
 \textsuperscript{}Oracle Health \& AI
\\
 \small{\{praphul.singh, charlotte.dzialo, jangwon.kim, sumana.srivatsa, irfan.bulu, sri.gadde, krishnaram.kenthapadi\}@oracle.com}
}

\begin{document}

\maketitle
\vspace{-3mm}
\begin{abstract}
Ensuring clinical data privacy while preserving utility is critical for AI-driven healthcare and data analytics. Existing de-identification (De-ID) methods, including rule-based techniques, deep learning models, and large language models (LLMs), often suffer from recall errors, limited generalization, and inefficiencies, limiting their real-world applicability. We propose a fully automated, multi-modal framework, \ourmodel\ for de-identifying structured and unstructured electronic health records, including clinical audio records. Our framework employs cost-efficient De-ID strategies, including intelligent routing, hybrid rule and LLM based approaches, and a two-step audio redaction approach. We present a retrieval-based entity relexicalization approach to ensure consistent substitutions of protected entities, thereby enhancing data coherence for downstream applications. We discuss key design desiderata, de-identification and relexicalization methodology, and modular architecture of \ourmodel\ and its integration with the Oracle Health Clinical AI system. Evaluated on the i2b2 2014 De-ID dataset using standard metrics with strict recall, our approach achieves competitive performance while optimizing token usage to reduce LLM costs. Finally, we discuss key lessons and insights from deployment in real-world AI-driven healthcare data pipelines.
\end{abstract}

\section{Introduction}
The proliferation of AI-driven healthcare tools has heightened the need for robust de-identification (De-ID) systems to comply with privacy regulations such as HIPAA in the US and GDPR in the EU~\cite{ahmed2020identification}. Effective De-ID is critical for secure AI model training, evaluation, and debugging, data analytics, and clinical deployment (see \S\ref{sec:intended_use_deid_data}). However, automating De-ID for electronic health records (EHRs) is challenging due to data heterogeneity, schema variability, context-sensitive Protected Health Information (PHI) or Personally Identifiable Information (PII), and the multi-modal nature of healthcare data—text, images, and audio~\cite{mohamed2023electronic, kayaalp2018patient}.

Manual De-ID, though accurate, is impractical at scale given the data volume in clinical settings~\cite{patterson2024call}. Automated approaches, including rule-based methods, BERT-based models, and LLMs~\cite{meystre2010automatic, kovavcevic2024identification, altalla2025evaluating}, face limitations in generalization, contextual reasoning, and efficiency, particularly when trained on narrow datasets that do not reflect real-world EHR diversity~\cite{liu2023deid}. Since even a single leak of PHI/PII can have serious privacy implications, reliable, scalable De-ID remains a critical need.

Recent advancements address cost, scalability, and generalizability through techniques like prompt optimization, model quantization~\cite{shekhar2024towards, arefeen2024leancontext}, and intelligent agent routing~\cite{varangot2025doing}, along with multi-modal De-ID for text and audio~\cite{dhingra2024speech}. Yet challenges persist -- specifically with, maintaining high recall, ensuring consistent PHI/PII substitution (e.g., “Wilson” and “Dr. Adam Wilson” both mapped to “Chang” and "Dr. Kevin Chang" respectively), and evaluating privacy risks with stricter metrics beyond token-level scores.

We propose \ourmodel, a fully automated, multi-modal framework for de-identifying structured and unstructured patient records, including clinical audio records. \ourmodel\ combines LLM-based processing for unstructured text with rule-based handling of structured data to achieve low cost and latency, and extends text de-identification for audio redaction. Our framework includes a novel retrieval-based entity relexicalization component to ensure consistent PHI/PII replacement, enhancing coherence and privacy. We present key design requirements (\S\ref{sec:design_desiderata}), de-identification and relexicalization methodology and architecture (\S\ref{sec:system}), and integration with the Oracle Health Clinical AI system (\S\ref{sec:system_integration}). We demonstrate that our framework outperforms other LLM-based approaches and achieves performance comparable to specialized, closed-source solutions while remaining adaptable through prompt engineering -- eliminating reliance on large annotated datasets (\S\ref{sec:experiments}), and highlight lessons and insights from 12+ months of deployment in real-world clinical AI system (\S\ref{sec:deployment_insights}).

\begin{figure*}[t]
    \centering
    \includegraphics[width=0.9\textwidth]{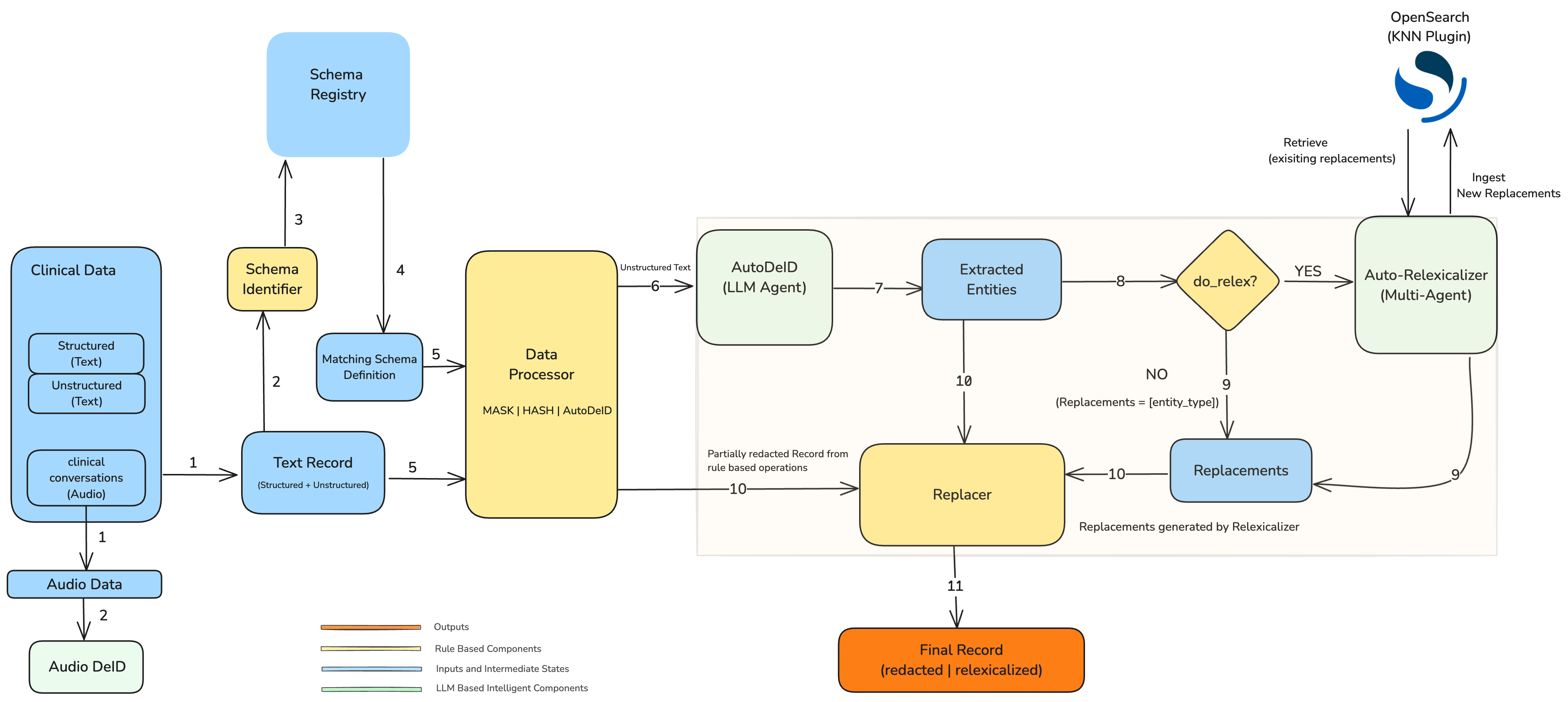}
    \caption{Architectural overview of \ourmodel}
    \label{fig:system_diagram}
\end{figure*}

\section{Related Work}
\label{sec:related_work}

{\noindent \bf Rule-based and ML-based De-ID}.
Rule-based systems rely on pattern matching, lexicons, and heuristics~\cite{neamatullah2008automated, meystre2010automatic}, offering simplicity and no training data requirements~\cite{negash2023identification}. However, rule creation is time-consuming and may lack robustness~\cite{negash2023identification,lee2017hybrid}. Machine learning models, especially BiLSTM-based approaches~\cite{ma2016end, dernoncourt2017neuroner, liu2017identification}, improve generalization without manual rules but struggle to transfer across datasets~\cite{stubbs2017identification, yang2019study}. BERT-based models enhance De-ID~\cite{meaney2022comparative} but demand significant compute resources, hyperparameter tuning, and still exhibit gaps in handling certain PHI/PII types.

{\noindent \bf LLM-based De-ID}. LLMs offer flexible, zero/few-shot De-ID capabilities. \citeauthor{kim2024generalizing} used GPT-4 to augment training data, improving BERT model performance across datasets. \citeauthor{yashwanth2024zero} showed fine-tuned LLMs outperform zero-shot models, particularly under format shifts. \citeauthor{altalla2025evaluating} found GPT-4 surpasses GPT-3.5 in De-ID accuracy and synthetic data generation. Similarly, \citeauthor{wiest2025deidentifying} developed a custom open-source LLM-based Anonymizer pipeline benchmarking 8 LLMs to De-ID 250 German clinical letters. However, most studies lack evaluation on large real-world cross-dataset generalization.

{\noindent \bf Synthetic Data}. The growing need for large datasets in medical research, alongside strict patient privacy rules, has led to increased interest in synthetic data. 
Synthetic data generation often involves differential privacy based approaches to protect patient privacy and generative adversarial networks (GAN) based methods for realistic data replication, and its utility depends on maintaining fidelity and minimizing biases to ensure reliable research and clinical decisions \cite{al2021differentially}. 
While synthetic data offers transformative potential for healthcare, careful consideration is needed to ensure its ethical and effective use in research and practice \cite{altalla2025evaluating}.

{\noindent \bf Relexicalization}. Replacing PHI/PII with realistic surrogates is underexplored. Many systems apply dummy replacements or simple rules (e.g., gender matching)~\cite{sweeney1996replacing, alfalahi2012pseudonymisation, lison2021anonymisation}. Recent work~\cite{vakili2024end} demonstrated pseudonymizing BERT models for privacy-preserving data analysis, highlighting relexicalization’s value in maintaining utility while protecting privacy.

\section{System Design and Architecture}
\label{sec:system}

\subsection{Design Desiderata}
\label{sec:design_desiderata}
\ourmodel\ is designed around three core principles: scalability, adaptability, and cost-efficiency. Scalability is achieved through an end-to-end automation pipeline, enabling efficient processing of structured and unstructured clinical data while minimizing computational overhead via intelligent routing and LLM-based De-ID strategies. Adaptability is ensured through a schema-agnostic processing architecture, facilitating seamless integration across heterogeneous EHR formats and multimodal data sources (text and audio) without the need for dataset-specific fine-tuning. Cost-efficiency is realized through token usage optimization in text-based De-ID and leveraging text-based entity extraction for audio, thereby eliminating the need for computationally expensive, audio-specific de-identification models. Additionally, retrieval-based re-lexicalization enhances contextual consistency in PHI/PII replacements, preserving both privacy and downstream utility, making the system highly effective for real-world AI-driven healthcare applications.

\subsection{Architecture Overview}
\ourmodel\ consists of three main components: (i) Auto De-ID, Audio De-ID, and Auto Relexicalizer (see Figure~\ref{fig:system_diagram}).


First, Schema Identifier automatically identifies the appropriate schema from the Schema Registry based on the \texttt{dataType} parameter in each data instance (see \S\ref{schema_definition_example}) and forwards it to the Data Processor along with the corresponding text data. The data processor is designed to be agnostic to text data types, requiring only the schema (stored in the schema registry) with predefined rule flags for each field. Currently, a rule flag can be one of the following: (i) \texttt{passThrough} (rule-based) retains the field without any changes (used for non-PHI/PII data), (ii) \texttt{shouldMask} (rule-based) replaces PHI/PII fields with generic placeholders (e.g., \texttt{[PERSON]}), (iii) \texttt{shouldHash} (rule-based pseudo-anonymization) hashes identifiers to enable secure linkages across documents within the same domain, or (iv) \texttt{autoDeID} (LLM-based) applies LLM-based De-ID to the unstructured text fields. This schema-agnostic design is crucial for scaling our system to support health data De-ID tasks. Meanwhile, audio data is processed separately by the Audio De-ID component. To ensure that no PHI/PII field definition is missed, we enforce a human review of the schema before it is pushed into the schema registry.

Auto De-ID is an LLM component (\S\ref{tab:deid_prompt_templates}) that processes the context extracted by the data processor. It can support a dynamic list of entity types. We support 33 entity types in our production deployment as shown in \S\ref{tab:entities}. This context is split into chunks of a pre-defined size ($\omega$), ensuring optimal model performance without exceeding LLM context length limits. Chunks are processed in parallel across a fixed number ($p$) of passes. 
$\omega$ and $p$ are heuristically chosen hyperparameters.
In each pass, the LLM extracts entities along with their surrounding context as position hints -- that is, each extracted entity includes nearby words that uniquely identify its location in the text (e.g., ``76 years old'' instead of just ``76'', or ``Mr. John Smith, the patient'' instead of just ``John''). This context-aware extraction enables accurate entity matching and redaction without relying on potentially unreliable character position indices.
In the first pass, the LLM detects as many entities as possible. In subsequent passes, entities already identified are masked in the text, prompting the model to focus on previously missed or hard-to-detect PHI/PII entities. Extracted entities from all passes are aggregated to form the final entity set. 

The Auto Relexicalizer, a multi-agent component (see \S\ref{tab:relex_prompt_templates} and Figure~\ref{fig:relex_workflow}), replaces redacted entities with contextually consistent and realistic alternatives. Relexicalization not only improves the usability of de-identified data but it also strengthens privacy by increasing the Hiding in Plain Sight (HIPS) factor \cite{carrell2020resilience}. Ensuring that replaced entities blend seamlessly with any remaining leaked PHI/PII makes re-identification attempts significantly more challenging. A combined example of Auto De-ID followed by Relexicalization is shown in \S\ref{tab:e2e_example}. It employs multiple agents as follows:
\begin{itemize}[noitemsep, topsep=0pt]
    \item LLM-Based Entity Clustering: grouping extracted entities based on their context.
    \item Hybrid Retrieval (Vector Search + Filtering): retrieving pre-existing replacements.
    \item LLM-Based Validation: Determining the validity of the retrieved replacements.
    \item LLM-Based Generation: Generating new replacements for invalid retrievals.
    \item OpenSearch Indexing: Storing new replacements for future reuse.
\end{itemize}

Our work extends a recent work~\cite{vakili2024end} that presents an analysis of pseudo-anonymization.  We offer an LLM-driven alternative for automated and scalable relexicalization. A regex-based replacer replaces extracted entities with entity-type masks (e.g., \texttt{[PERSON]}) to redact PHI/PII in unstructured text fields.
See \S\ref{sec:app_auto_deid_example} for an end-to-end example of Auto De-ID.

Our Audio De-ID feature performs a two-step redaction process to enhance privacy.  First, it uses Automatic Speech Recognition (ASR) to detect timestamps of spoken words and applies LLM-based Auto De-ID to the transcript, adding an extra 100 -- 200 msec of margin at token boundaries for improved protection. Next, it examines \textit{unrecognized} voiced regions -- identified by an aggressive Voice Activity Detection (VAD) -- by analyzing their surrounding context words with an LLM (\S\ref{tab:audio_prompt_templates}), evaluating the likelihood of these regions containing PHI/PII and selects the most likely ones.  Finally, it mutes all time boundaries (including margins) for predicted PHI/PII tokens' voiced regions.  Testing on our internal data showed that the second step increased recall by approximately 10\%. A brief end-to-end example illustrating this is presented in \S\ref{sec:app_audio_deid_example}.  In summary, our Audio De-ID component improves PHI/PII detection by addressing ASR misalignment and deletion errors by leveraging VAD and LLM based detection process in the second step.

By integrating Auto De-ID for unstructured text, Auto Relexicalizer for realistic entity replacement, and Audio De-ID for speech data, \ourmodel\ provides a scalable, adaptable, and cost-effective De-ID pipeline that secures both text and audio data while preserving its utility.

See \S\ref{sec:app_autodeid_algo}, \S\ref{sec:app_audio_deid}, and \S\ref{sec:app_auto_relex} for detailed algorithmic descriptions of Auto De-ID, Auto Relexicalization, and Audio De-ID, respectively.

\section{Experiments}
\label{sec:experiments}
We present the results of evaluating \ourmodel\ against other LLM-based approaches and specialized, closed-source commercial solutions over a publicly available medical record dataset.
For parity with other methods, we turn off the Auto Relexicalizer component. We set chunk size ($\omega$) to 256 and number of passes ($p$) to 2.

\subsection{Dataset}
We evaluated using 2014 i2b2/UTHealth De-ID corpus~\cite{stubbs2015annotating} which is widely used in clinical De-ID research. This dataset comprises longitudinal clinical records for 296 patients (with 2-5 records per patient). The annotation scheme follows HIPAA guidelines and includes additional indirect identifiers such as detailed date components (e.g., year), geographic information (states, countries), hospital names, clinician names, and patient professions. For our experiments, we randomly subsampled 100 clinical notes and evaluated on seven PHI/PII entity categories: AGE, CONTACT, DATE, ID, LOCATION, NAME, and PROFESSION.

\subsection{Comparative De-ID methods}
We evaluated \ourmodel\ against recent LLM-based methods, \citet{yashwanth2024zero} (with their two prompt variants `brief' and `detailed') and  \citet{altalla2025evaluating}, as well as commercial De-ID APIs from AWS (Amazon Web Services) \cite{aws_comprehend_medical} and JSL (John Snow Labs) \cite{kocaman2023rwd143, kocaman2025can}. For a fair comparison, we used GPT-4o \cite{openai_gpt4o} for all LLM-based methods. See \S\ref{sec:app_other_models} for additional details.

\subsection{Evaluation Methods}
We assessed De-ID performance using traditional metrics such as precision, recall and F1-score, as well as all-or-nothing recall was applied to (PERSON, AGE, CONTACT, ID, LOCATION). All-or-nothing recall \cite{scaiano2016unified} determines whether every instance of a given entity type or a document is correctly redacted. If any instance is missed, all-or-nothing recall is set to 0; otherwise, it is set to 1.

We evaluated all systems using a \textit{stricter} methodology for true positive computation incorporating entity position matching. Position information is critical for data redaction in unstructured health records, as it often differentiates PHI from clinical information. For example, in the phrase `76 yrs old,' the number `76' represents age (PHI), whereas in `oxygen saturation rate is 76,' it denotes a vital sign.

In the evaluation for each entity type, we compared our system with AWS and JSL using additional matching criteria, including entity-level text matching and label matching. The metrics were assessed at the PHI/PII entity level (multi-word spans) rather than individual tokens, as in \cite{yashwanth2024zero}, ensuring a fair comparison between entities with varying token counts. To account for minor variations (e.g., `Mrs. Mary Smith' vs. `Mary Smith'), we applied the Levenshtein similarity with a heuristically determined threshold of 0.6. \citet{yashwanth2024zero} and \citet{altalla2025evaluating} are omitted in entity type specific evaluation, because they provide only binary PHI/PII labels, but not entity types.

\begin{table}[t]
    \centering
    \small
    \begin{tabular}{l c c c p{1.1cm}}
        \hline
        \textbf{Model} & \textbf{Precision} & \textbf{Recall} & \textbf{F1-score} & \textbf{All-Or-Nothing Recall} \\ 
        \hline
        Y\&S\_Brief & 0.5634 & 0.6580 & 0.6070 & 0.3700 \\
        Y\&S\_Detail & 0.6178 & 0.8270 & 0.7072 & 0.5600 \\
        Altalla & 0.9675 & 0.6715 & 0.7927 & 0.3600 \\
        \ourmodel\ & 0.9769 & 0.9525 & 0.9646 & 0.7900 \\
        \hline
        
        AWS & 0.9549 & 0.9425 & 0.9487 & 0.7500 \\
        JSL & 0.9481 & 0.9865 & 0.9669 & 0.9000 \\
        \hline
    \end{tabular}
    \caption{Performance of zero-shot GPT-4o and commercial De-ID systems on all PHI/PII entities. This evaluation does not consider the entity type constraint.}
    \label{tab:doc_level_metric}
\end{table}

\begin{table*}[t]
    \centering
    \small
    \renewcommand{\arraystretch}{1.2}
    \begin{tabular}{lccccccccc}
        \hline
        Entity Type & \multicolumn{3}{c|}{\ourmodel} & \multicolumn{3}{c|}{AWS} & \multicolumn{3}{c}{JSL} \\
        \cline{2-10}
         & Recall & Precision & F1-score & Recall & Precision & F1-score & Recall & Precision & F1-score \\
        \hline
        AGE & 0.8987 & 0.9930 & 0.9435 & 0.9684 & 0.9935 & 0.9808 & 0.9748 & 0.9688 & 0.9718 \\
        CONTACT & 1.0000 & 1.0000 & 1.0000 & 1.0000 & 0.4545 & 0.6250 & 0.7879 & 1.0000 & 0.8814 \\
        DATE & 0.9495 & 0.9988 & 0.9735 & 0.9202 & 0.9949 & 0.9561 & 0.9918 & 0.9883 & 0.9900 \\
        ID & 0.9275 & 0.6667 & 0.7758 & 0.7917 & 0.6129 & 0.6909 & 0.8667 & 0.9123 & 0.8889 \\
        LOCATION & 0.7855 & 1.0000 & 0.8799 & 0.7890 & 0.9820 & 0.8750 & 0.9469 & 0.9808 & 0.9636 \\
        PERSON & 0.9595 & 0.9912 & 0.9751 & 0.9461 & 0.9461 & 0.9461 & 0.9572 & 0.9933 & 0.9749 \\
        PROFESSION & 0.9167 & 1.0000 & 0.9565 & 0.9130 & 0.7778 & 0.8400 & 1.0000 & 0.9565 & 0.9778 \\
        \hline
        All & 0.9159 & 0.9790 & 0.9465 & 0.9042 & 0.9510 & 0.9270 & 0.9664 & 0.9839 & 0.9751 \\
        \hline
    \end{tabular}
    \caption{Performance of De-ID systems for each entity type and all data.}
    \label{tab:entity_deid_performance}
\end{table*}

\subsection{Comparison with LLM-Based De-ID Frameworks}
As shown in Table~\ref{tab:doc_level_metric}, a comparison with other LLM-based methods \cite{yashwanth2024zero, altalla2025evaluating} indicates that \ourmodel\ outperforms existing methods, achieving the highest F1-score of 0.9646 with a well-balanced precision and recall. 
The high recall can be attributed to \ourmodel’s multi-chunk and multi-pass strategy, which systematically refines entity detection by iteratively masking extracted entities and forcing the model to focus on overlooked PHI. Similarly, in terms of precision, \ourmodel\ outperforms \citet{yashwanth2024zero}, highlighting the effectiveness of its context-aware entity extraction. By leveraging contextual clues and maintaining intra-document consistency, \ourmodel\ reduces false positives, whereas single-pass prompting methods tend to over-redact ambiguous terms.

While \citet{yashwanth2024zero}'s Detailed version achieves higher recall than the Brief, it does so at the expense of precision. This highlights a fundamental trade-off in zero-shot protected terms extraction: models optimized for recall often over-mask non-protected terms, leading to reduced utility of the redacted text. \ourmodel\ strikes a balance between recall and precision, making it more suitable for real-world clinical applications where both PHI/PII removal and utility are essential.

\subsection{Comparisons with Commercial APIs}
Unlike prior zero-shot LLM-based approaches, commercial De-ID APIs (e.g., AWS, JSL) are fine-tuned on proprietary clinical datasets. Table~\ref{tab:doc_level_metric} shows that while \ourmodel\ does not surpass JSL in recall, it achieves higher precision and a comparable F1-score. This suggests that while JSL benefits from domain-specific fine-tuning, \ourmodel’s context-aware extraction minimizes false positives, leading to more reliable entity masking.

\begin{table}[b]
    \centering
    \small
    \begin{tabular}{l c c c c}
        \hline
        {Entity Type} & {\ourmodel} & {AWS} & {JSL} \\ 
        \hline
        AGE & 0.8904 & 0.9589 & 0.9452 \\
        CONTACT & 1.000 & 1.0000 & 0.7666 \\
        DATE & 0.7300 & 0.6000 & 0.9500  \\
        ID & 0.8958 & 0.8333 & 0.8541 \\
        LOCATION & 0.6923 & 0.6026 & 0.8333 \\
        PERSON & 0.8556 & 0.8041 & 0.8659 \\
        PROFESSION & 0.9091 & 0.9090 & 1.0000 \\
        \hline
        All & 0.8214 & 0.7701 & 0.8906 \\
        \hline
    \end{tabular}
    \caption{All-or-nothing recalls constrained on entity types for \ourmodel\ and commercial models.}
    \label{tab:all_or_nothing_recall_entity_type}
\end{table}

Table~\ref{tab:entity_deid_performance} presents a breakdown of performance by entity types. \ourmodel\ demonstrates high precision and strong recall across all entities, particularly excelling on CONTACT and PERSON entities. \ourmodel\ achieves perfect recall on CONTACT entities, outperforming AWS and JSL, and shows competitive performance on PERSON and DATE entities. However, it underperforms on LOCATION and ID, presumably due to the structural variability of PHI in clinical texts. LOCATION entities, in particular, often appear within complex sentence structures, posing challenges to generic LLM-based masking. This suggests the need for instruction updates for these entities or a specialized LLM.

\subsection{All-or-Nothing Recall Results}
Table~\ref{tab:all_or_nothing_recall_entity_type} shows performance by PHI/PII entity type with all-or-nothing recall, a stricter metric requiring both correct entity type and position alignment. These results highlight \ourmodel’s strength in high-precision redaction for certain entities while emphasizing the advantage of domain-tuned models for broader recall coverage. 

\ourmodel\ outperforms other methods using the same underlying LLM (Table~\ref{tab:doc_level_metric}) and approaches the performance of specialized models like JSL. Specifically, it excels in CONTACT and ID compared to commercial methods. \ourmodel's multi-pass extraction and context-aware masking enhance the LLM's effectiveness, showcasing its strength without specialized fine-tuning. However, JSL leads in most other entity types, achieving the highest overall recall. The recall gap -- especially for complex entities like DATE and LOCATION -- highlights the need for improved prompt instructions and possibly a specialized LLM for de-identification to match domain-specific systems.

\subsection{Ablation Study with Open-Source Model}
\label{sec:ablation}

To demonstrate the adaptability of \ourmodel\ to open-source models and evaluate the benefit of its multi-pass de-identification strategy, we conducted an ablation study using LLaMA-3.2-3B-Instruct \cite{meta2024llama} -- a compact, publicly available LLM.

\begin{figure}[t]
    \centering
    \includegraphics[width=0.95\linewidth]{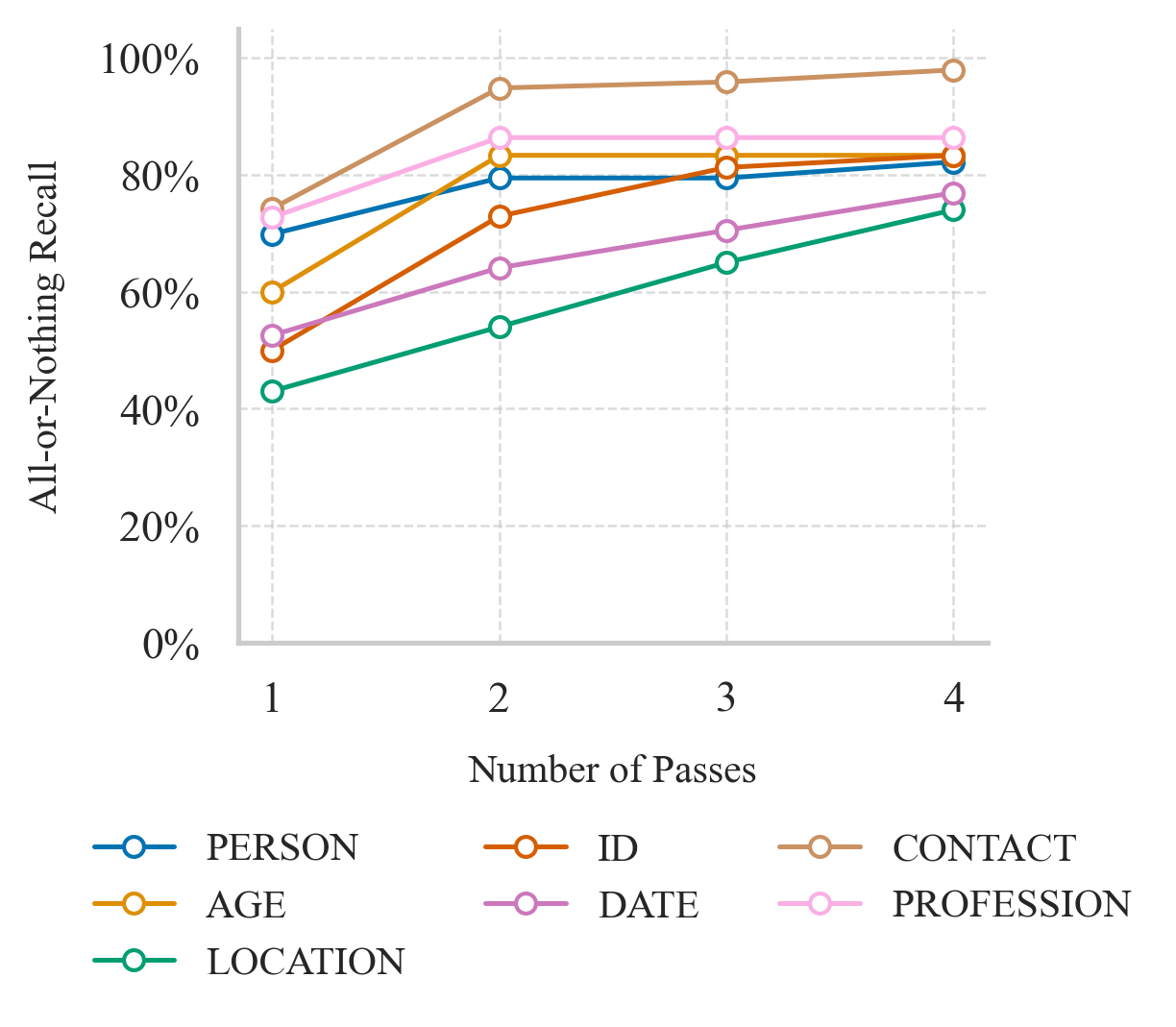}
    \caption{Entity-wise all-or-nothing recall for LLaMA-3.2-3B as the number of passes increases from 1 to 4. Most entities show the greatest gain between pass 1 and 2, with diminishing improvements thereafter.}
    \label{fig:ablation_llama3.2}
\end{figure}

Figure~\ref{fig:ablation_llama3.2} illustrates the effect of increasing the number of passes (from 1 to 4) on all-or-nothing recall across seven PHI/PII entity types. Notably, we observe consistent improvement across most entity types as the number of passes increases. The largest relative gains are observed between pass 1 and pass 2, especially for sparse or context-sensitive types such as \texttt{ID}, \texttt{DATE}, and \texttt{LOCATION}, which tend to be missed in early passes but are recovered in subsequent iterations.

For dominant or well-signaled types like \texttt{PERSON}, \texttt{CONTACT}, and \texttt{PROFESSION}, the recall is already high at pass 2, with marginal improvements beyond that point. By pass 3, the recall curve starts to saturate for most entity types, indicating diminishing returns on additional passes.

These trends suggest that the number of passes is a critical, model-dependent hyperparameter: lightweight models like LLaMA-3.2-3B benefit from 2–3 passes, while larger models may reach optimal performance sooner. \ourmodel\ supports this flexibility by treating the pass count as a configurable parameter, allowing practitioners to trade off between computational cost and de-identification completeness depending on the capacity of the underlying LLM.

\subsection{Qualitative Evaluation of Audio De-ID on Internal Data}

To assess the impact of our two-step Audio De-ID process, we conducted a qualitative evaluation on an internal clinical audio dataset. The LLM-based timestamp detector identified and muted several additional audio segments that were missed by text de-identification on just the transcript. A subset of these newly muted segments included person names that had previously gone undetected, leading to a noticeable improvement in all-or-nothing recall on direct identifiers—approximately 12\%. Another small portion involved relevant medical or personal context (e.g., complaints or medications), introducing minor precision trade-offs. The majority of muted segments, however, were non-informative, consisting of background noise or idle speech such as keyboard activity. Overall, over 84\% of the additionally muted content was deemed to have no negative impact on clinical utility. These findings demonstrate that the second-pass audio detection enhances recall with minimal utility loss, validating its inclusion in real-world deployments.

\section{Deployment Lessons and Insights}\label{sec:deployment_insights}
In the course of developing and deploying \ourmodel, we realized that efficient scaling of de-identification is crucial to handle the large volume of healthcare data post-deployment, including audio files, SOAP notes, and longitudinal records with thousands of FHIR 
resources \cite{bender2013hl7}.  Once the service is launched, a continuous influx of data follows, and as the products expand, the ability to optimize processing at scale becomes critical. Simple yet powerful reductions in computation and processing can significantly impact efficiency, cost, and system performance.

One key optimization we employed was reducing token usage. \ourmodel\ extracts only PHI/PII entities and their position hints, minimizing the number of LLM's output tokens by approximately 50\%. This reduction not only lowers processing costs but also decreases latency, ensuring that De-ID remains accurate and efficient at scale.

Further, processing each FHIR resource individually causes delays and backlogs. Our schema agnostic approach allows us to batch lightweight resources (e.g., vitals, medications) in the schema processor by merging their schemas and free-text fields into a single composite schema. For example, if the batch size is $n$, $n$ resource schemas could be combined into one, allowing all associated text to be de-identified in a single LLM request. The batch size can be chosen heuristically based on the LLM used, the context length it supports, the system prompt's size, and the average number of tokens present in the unstructured texts of the resource schemas. 

Finally, dynamic batching further enhances scalability by grouping incoming resources based on size and complexity.  This approach enables large and diverse datasets to be processed in real time, preventing bottlenecks as data streams grow.

Our initial implementation focused on just de-identification but we decided to incorporate relexicalization after realizing that relexicalized data significantly enhances the ease of use by applied scientists as part of their model training, quality \& bias evaluation, and debugging pipelines since this data has similar format and characteristics as the production data.

\section{Conclusion And Future Directions}\label{sec:conclusion}
Motivated by the need for protecting patient privacy while enabling utility, we presented \ourmodel\, a multi-modal, scalable, flexible, and cost-efficient LLM-powered framework for clinical data de-identification, and demonstrated its efficacy in de-identifying 33 PHI/PII entities over the i2b2 dataset. We showed that our approach outperforms other LLM-based methods and achieves performance comparable to specialized, closed-source solutions. Further, \ourmodel\ supports relexicalizing redacted entities with contextually consistent alternatives, enhancing data usability and strengthening privacy. By presenting the methodology, technical architecture, and lessons learned from over 12 months of production deployment, we hope that the insights and experience from our work are useful for researchers and practitioners working on clinical AI systems.

There are several avenues for future work. 
The variability in healthcare datasets across institutions affects generalizability, necessitating adaptive prompting techniques. While our method excels in detecting \texttt{ID}, \texttt{PERSON}, and \texttt{DATE} entities, it may require further refinement of entity-specific LLM instructions (e.g., for address- and occupation-related entities). More broadly, \ourmodel’s generalizability can be enhanced across diverse institutional datasets by integrating domain-adaptive prompt enhancements.

Another direction is to investigate domain-adaptive VAD techniques in de-identification settings. 
Although we incorporate a VAD-based solution to mitigate ASR inaccuracies, our use of a simpler VAD algorithm introduces false positives, leading to over-redaction. Additionally, transcription variability due to noise and overlapping speech increases the risk of PHI/PII leakage. Integrating deep learning-based VAD models alongside domain-adaptive ASR techniques could enhance precision while maintaining recall, reducing unnecessary redactions without compromising PHI/PII protection.

Furthermore, we could design standardized benchmarks for evaluating relexicalization techniques, following ideas discussed in \S\ref{tab:relex_metrics}. An ideal dataset would include PHI/PII-tagged text, gold relexicalized outputs, and context to ensure consistency across documents and domains such as clinical notes, transcripts, and structured records. More broadly, a promising direction is to extend our framework to handle other modalities such as medical images and videos, and design corresponding end-to-end evaluation methodolgies and benchmarks.

\section*{Acknowledgments}
We would like to thank other members of Oracle Health AI for their collaboration while deploying \ourmodel\ in production, and 
Weifeng Bao,
Brent Beardsley,
Michelle Chen,
Dipankar Das,
Long Duong,
Raefer Gabriel,
Kent Grueneich,
Neil Hauge,
Bhagya Hettige,
Brad Jacobs,
Mark Johnson,
Shirley Liu,
Cody Maheu,
Atri Mandal,
Virendra Marathe,
Kiran Rama,
Amitabh Saikia,
Tushar Shandhilya,
Gyan Shankar,
and
Vishal Vishnoi
for insightful feedback and discussions.

\bibliography{ohai}

\begin{thebibliography}{38}
\providecommand{\natexlab}[1]{#1}

\bibitem[{Ahmed et~al.(2020)Ahmed, Aziz, and Mohammed}]{ahmed2020identification}
Tanbir Ahmed, Md~Momin~Al Aziz, and Noman Mohammed. 2020.
\newblock De-identification of electronic health record using neural network.
\newblock \emph{Scientific reports}, 10(1):18600.

\bibitem[{Al~Aziz et~al.(2021)Al~Aziz, Ahmed, Faequa, Jiang, Yao, and Mohammed}]{al2021differentially}
Md~Momin Al~Aziz, Tanbir Ahmed, Tasnia Faequa, Xiaoqian Jiang, Yiyu Yao, and Noman Mohammed. 2021.
\newblock Differentially private medical texts generation using generative neural networks.
\newblock \emph{ACM Transactions on Computing for Healthcare (HEALTH)}, 3(1):1--27.

\bibitem[{Alfalahi et~al.(2012)Alfalahi, Brissman, and Dalianis}]{alfalahi2012pseudonymisation}
Alyaa Alfalahi, Sara Brissman, and Hercules Dalianis. 2012.
\newblock Pseudonymisation of personal names and other {PHIs} in an annotated clinical {Swedish} corpus.
\newblock In \emph{Third Workshop on Building and Evaluating Resources for Biomedical Text Mining (BioTxtM 2012) Held in Conjunction with LREC}, pages 49--54.

\bibitem[{Altalla’ et~al.(2025)Altalla’, Abdalla, Altamimi, Bitar, Al~Omari, Kardan, and Sultan}]{altalla2025evaluating}
Bayan Altalla’, Sameera Abdalla, Ahmad Altamimi, Layla Bitar, Amal Al~Omari, Ramiz Kardan, and Iyad Sultan. 2025.
\newblock Evaluating {GPT} models for clinical note de-identification.
\newblock \emph{Scientific Reports}, 15(1):3852.

\bibitem[{Arefeen et~al.(2024)Arefeen, Debnath, and Chakradhar}]{arefeen2024leancontext}
Md~Adnan Arefeen, Biplob Debnath, and Srimat Chakradhar. 2024.
\newblock {LeanContext}: Cost-efficient domain-specific question answering using {LLMs}.
\newblock \emph{Natural Language Processing Journal}, 7:100065.

\bibitem[{AWS(2025)}]{aws_comprehend_medical}
AWS. 2025.
\newblock {Amazon Comprehend Medical}.
\newblock \url{https://aws.amazon.com/comprehend/medical/}.
\newblock Accessed: March, 2025.

\bibitem[{Bender and Sartipi(2013)}]{bender2013hl7}
Duane Bender and Kamran Sartipi. 2013.
\newblock {HL7 FHIR}: An agile and {RESTful} approach to healthcare information exchange.
\newblock In \emph{Proceedings of the 26th IEEE international symposium on computer-based medical systems}, pages 326--331. IEEE.

\bibitem[{Carrell et~al.(2020)Carrell, Malin, Cronkite, Aberdeen, Clark, Li, Bastakoty, Nyemba, and Hirschman}]{carrell2020resilience}
David~S Carrell, Bradley~A Malin, David~J Cronkite, John~S Aberdeen, Cheryl Clark, Muqun Li, Dikshya Bastakoty, Steve Nyemba, and Lynette Hirschman. 2020.
\newblock Resilience of clinical text de-identified with “hiding in plain sight” to hostile reidentification attacks by human readers.
\newblock \emph{Journal of the American Medical Informatics Association}, 27(9):1374--1382.

\bibitem[{Dernoncourt et~al.(2017)Dernoncourt, Lee, and Szolovits}]{dernoncourt2017neuroner}
Franck Dernoncourt, Ji~Young Lee, and Peter Szolovits. 2017.
\newblock {NeuroNER}: An easy-to-use program for named-entity recognition based on neural networks.
\newblock \emph{arXiv preprint arXiv:1705.05487}.

\bibitem[{Dhingra et~al.(2024)Dhingra, Agrawal, Veerappan, Ho, Chng, and Tong}]{dhingra2024speech}
Priyanshu Dhingra, Satyam Agrawal, Chandra~Sekar Veerappan, Thi~Nga Ho, Eng~Siong Chng, and Rong Tong. 2024.
\newblock Speech de-identification data augmentation leveraging large language model.
\newblock In \emph{2024 International Conference on Asian Language Processing (IALP)}, pages 97--102. IEEE.

\bibitem[{Kayaalp(2018)}]{kayaalp2018patient}
Mehmet Kayaalp. 2018.
\newblock Patient privacy in the era of big data.
\newblock \emph{Balkan medical journal}, 35(1):8--17.

\bibitem[{Kim et~al.(2024)Kim, Hahm, and Lee}]{kim2024generalizing}
Woojin Kim, Sungeun Hahm, and Jaejin Lee. 2024.
\newblock Generalizing clinical de-identification models by privacy-safe data augmentation using {GPT}-4.
\newblock In \emph{Proceedings of the 2024 Conference on Empirical Methods in Natural Language Processing}, pages 21204--21218.

\bibitem[{Kocaman et~al.(2025)Kocaman, Santas, Gul, Butgul, and Talby}]{kocaman2025can}
Veysel Kocaman, Muhammed Santas, Yigit Gul, Mehmet Butgul, and David Talby. 2025.
\newblock Can zero-shot commercial {APIs} deliver regulatory-grade clinical text deidentification?
\newblock In \emph{ECIR Workshop on Narrative Extraction from Texts (Text2Story)}.

\bibitem[{Kocaman et~al.(2023)Kocaman, Talby, and Hak}]{kocaman2023rwd143}
Veysel Kocaman, D~Talby, and H~Ul Hak. 2023.
\newblock {RWD143} beyond accuracy: Automated de-identification of large real-world clinical text datasets.
\newblock \emph{Value in Health}, 26(12):S532.

\bibitem[{Kova{\v{c}}evi{\'c} et~al.(2024)Kova{\v{c}}evi{\'c}, Ba{\v{s}}aragin, Milo{\v{s}}evi{\'c}, and Nenadi{\'c}}]{kovavcevic2024identification}
Aleksandar Kova{\v{c}}evi{\'c}, Bojana Ba{\v{s}}aragin, Nikola Milo{\v{s}}evi{\'c}, and Goran Nenadi{\'c}. 2024.
\newblock De-identification of clinical free text using natural language processing: A systematic review of current approaches.
\newblock \emph{Artificial intelligence in medicine}, page 102845.

\bibitem[{Lee et~al.(2017)Lee, Wu, Zhang, Xu, Xu, and Roberts}]{lee2017hybrid}
Hee-Jin Lee, Yonghui Wu, Yaoyun Zhang, Jun Xu, Hua Xu, and Kirk Roberts. 2017.
\newblock A hybrid approach to automatic de-identification of psychiatric notes.
\newblock \emph{Journal of biomedical informatics}, 75:S19--S27.

\bibitem[{Lison et~al.(2021)Lison, Pil{\'a}n, S{\'a}nchez, Batet, and {\O}vrelid}]{lison2021anonymisation}
Pierre Lison, Ildik{\'o} Pil{\'a}n, David S{\'a}nchez, Montserrat Batet, and Lilja {\O}vrelid. 2021.
\newblock Anonymisation models for text data: State of the art, challenges and future directions.
\newblock In \emph{Proceedings of the 59th Annual Meeting of the Association for Computational Linguistics and the 11th International Joint Conference on Natural Language Processing (Volume 1: Long Papers)}, pages 4188--4203.

\bibitem[{Liu et~al.(2017)Liu, Tang, Wang, and Chen}]{liu2017identification}
Zengjian Liu, Buzhou Tang, Xiaolong Wang, and Qingcai Chen. 2017.
\newblock De-identification of clinical notes via recurrent neural network and conditional random field.
\newblock \emph{Journal of biomedical informatics}, 75:S34--S42.

\bibitem[{Liu et~al.(2023)Liu, Huang, Yu, Zhang, Wu, Cao, Dai, Zhao, Li, Shu et~al.}]{liu2023deid}
Zhengliang Liu, Yue Huang, Xiaowei Yu, Lu~Zhang, Zihao Wu, Chao Cao, Haixing Dai, Lin Zhao, Yiwei Li, Peng Shu, et~al. 2023.
\newblock {DeID-GPT}: Zero-shot medical text de-identification by {GPT}-4.
\newblock \emph{arXiv preprint arXiv:2303.11032}.

\bibitem[{Ma and Hovy(2016)}]{ma2016end}
Xuezhe Ma and Eduard Hovy. 2016.
\newblock End-to-end sequence labeling via bi-directional lstm-cnns-crf.
\newblock \emph{arXiv preprint arXiv:1603.01354}.

\bibitem[{Meaney et~al.(2022)Meaney, Hakimpour, Kalia, and Moineddin}]{meaney2022comparative}
Christopher Meaney, Wali Hakimpour, Sumeet Kalia, and Rahim Moineddin. 2022.
\newblock A comparative evaluation of transformer models for de-identification of clinical text data.
\newblock \emph{arXiv preprint arXiv:2204.07056}.

\bibitem[{MetaAI(2024)}]{meta2024llama}
MetaAI. 2024.
\newblock \href {https://ai.meta.com/blog/llama-3-2-connect-2024-vision-edge-mobile-devices/} {{LLaMA 3.2}}.
\newblock Accessed: 2025-05-20.

\bibitem[{Meystre et~al.(2010)Meystre, Friedlin, South, Shen, and Samore}]{meystre2010automatic}
Stephane~M Meystre, F~Jeffrey Friedlin, Brett~R South, Shuying Shen, and Matthew~H Samore. 2010.
\newblock Automatic de-identification of textual documents in the electronic health record: A review of recent research.
\newblock \emph{BMC medical research methodology}, 10:1--16.

\bibitem[{Mohamed et~al.(2023)Mohamed, Song, McMahon, Sahil, Zozus, Wang, Collaborative, and Waitman}]{mohamed2023electronic}
Yahia Mohamed, Xing Song, Tamara~M McMahon, Suman Sahil, Meredith Zozus, Zhan Wang, Greater~Plains Collaborative, and Lemuel~R Waitman. 2023.
\newblock Electronic health record data quality variability across a multistate clinical research network.
\newblock \emph{Journal of Clinical and Translational Science}, 7(1):e130.

\bibitem[{Neamatullah et~al.(2008)Neamatullah, Douglass, Lehman, Reisner, Villarroel, Long, Szolovits, Moody, Mark, and Clifford}]{neamatullah2008automated}
Ishna Neamatullah, Margaret~M Douglass, Li-Wei~H Lehman, Andrew Reisner, Mauricio Villarroel, William~J Long, Peter Szolovits, George~B Moody, Roger~G Mark, and Gari~D Clifford. 2008.
\newblock Automated de-identification of free-text medical records.
\newblock \emph{BMC medical informatics and decision making}, 8:1--17.

\bibitem[{Negash et~al.(2023)Negash, Katz, Neilson, Moni, Nesca, Singer, and Enns}]{negash2023identification}
Bekelu Negash, Alan Katz, Christine~J Neilson, Moniruzzaman Moni, Marcello Nesca, Alexander Singer, and Jennifer~E Enns. 2023.
\newblock De-identification of free text data containing personal health information: A scoping review of reviews.
\newblock \emph{International Journal of Population Data Science}, 8(1):2153.

\bibitem[{OpenAI(2025)}]{openai_gpt4o}
OpenAI. 2025.
\newblock {GPT-4o} documentation.
\newblock \url{https://platform.openai.com/docs/models/gpt-4o}.
\newblock Accessed: March, 2025.

\bibitem[{Patterson et~al.(2024)Patterson, Hekman, Liao, Hamedani, Shah, and Afshar}]{patterson2024call}
Brian~W Patterson, Daniel~J Hekman, Frank~J Liao, Azita~G Hamedani, Manish~N Shah, and Majid Afshar. 2024.
\newblock Call me {Dr Ishmael}: Trends in electronic health record notes available at emergency department visits and admissions.
\newblock \emph{JAMIA open}, 7(2):ooae039.

\bibitem[{Scaiano et~al.(2016)Scaiano, Middleton, Arbuckle, Kolhatkar, Peyton, Dowling, Gipson, and El~Emam}]{scaiano2016unified}
Martin Scaiano, Grant Middleton, Luk Arbuckle, Varada Kolhatkar, Liam Peyton, Moira Dowling, Debbie~S Gipson, and Khaled El~Emam. 2016.
\newblock A unified framework for evaluating the risk of re-identification of text de-identification tools.
\newblock \emph{Journal of biomedical informatics}, 63:174--183.

\bibitem[{Shekhar et~al.(2024)Shekhar, Dubey, Mukherjee, Saxena, Tyagi, and Kotla}]{shekhar2024towards}
Shivanshu Shekhar, Tanishq Dubey, Koyel Mukherjee, Apoorv Saxena, Atharv Tyagi, and Nishanth Kotla. 2024.
\newblock Towards optimizing the costs of {LLM} usage.
\newblock \emph{arXiv preprint arXiv:2402.01742}.

\bibitem[{Stubbs et~al.(2017)Stubbs, Filannino, and Uzuner}]{stubbs2017identification}
Amber Stubbs, Michele Filannino, and {\"O}zlem Uzuner. 2017.
\newblock De-identification of psychiatric intake records: Overview of {2016 CEGS N-GRID} shared tasks {Track} 1.
\newblock \emph{Journal of biomedical informatics}, 75:S4--S18.

\bibitem[{Stubbs and Uzuner(2015)}]{stubbs2015annotating}
Amber Stubbs and {\"O}zlem Uzuner. 2015.
\newblock Annotating longitudinal clinical narratives for de-identification: The 2014 {i2b2/UTHealth} corpus.
\newblock \emph{Journal of biomedical informatics}, 58:S20--S29.

\bibitem[{Sweeney(1996)}]{sweeney1996replacing}
Latanya Sweeney. 1996.
\newblock Replacing personally-identifying information in medical records, the {Scrub} system.
\newblock In \emph{Proceedings of the AMIA annual fall symposium}, page 333.

\bibitem[{Vakili et~al.(2024)Vakili, Henriksson, and Dalianis}]{vakili2024end}
Thomas Vakili, Aron Henriksson, and Hercules Dalianis. 2024.
\newblock End-to-end pseudonymization of fine-tuned clinical {BERT} models: Privacy preservation with maintained data utility.
\newblock \emph{BMC Medical Informatics and Decision Making}, 24(1):162.

\bibitem[{Varangot-Reille et~al.(2025)Varangot-Reille, Bouvard, Gourru, Ciancone, Schaeffer, and Jacquenet}]{varangot2025doing}
Clovis Varangot-Reille, Christophe Bouvard, Antoine Gourru, Mathieu Ciancone, Marion Schaeffer, and Fran{\c{c}}ois Jacquenet. 2025.
\newblock Doing more with less--implementing routing strategies in large language model-based systems: An extended survey.
\newblock \emph{arXiv preprint arXiv:2502.00409}.

\bibitem[{Wiest et~al.(2025)Wiest, Le{\ss}mann, Wolf, Ferber, Treeck, Zhu, Ebert, Westphalen, Wermke, and Kather}]{wiest2025deidentifying}
Isabella~C Wiest, Marie-Elisabeth Le{\ss}mann, Fabian Wolf, Dyke Ferber, Marko~Van Treeck, Jiefu Zhu, Matthias~P Ebert, Christoph~Benedikt Westphalen, Martin Wermke, and Jakob~Nikolas Kather. 2025.
\newblock Deidentifying medical documents with local, privacy-preserving large language models: The {LLM}-anonymizer.
\newblock \emph{NEJM AI}, 2(4):AIdbp2400537.

\bibitem[{Yang et~al.(2019)Yang, Lyu, Li, Lee, Bian, Hogan, and Wu}]{yang2019study}
Xi~Yang, Tianchen Lyu, Qian Li, Chih-Yin Lee, Jiang Bian, William~R Hogan, and Yonghui Wu. 2019.
\newblock A study of deep learning methods for de-identification of clinical notes in cross-institute settings.
\newblock \emph{BMC medical informatics and decision making}, 19:1--9.

\bibitem[{Yashwanth and Shettar(2024)}]{yashwanth2024zero}
YS~Yashwanth and Rajashree Shettar. 2024.
\newblock Zero and few short learning using large language models for de-identification of medical records.
\newblock \emph{IEEE Access}.

\end{thebibliography}

\appendix

\section{Appendix}

\subsection{Ethics Statement}
Despite the high performance of our De-ID system, there remains a non-zero risk that some PHI/PII might not be detected or removed. Consequently, any output produced by automated De-ID system should still be handled with the same security and privacy precautions as raw identifiable data.  We underscore that users of our De-ID framework should apply rigorous privacy safeguards when handling the processed data, just as they would for original clinical records. For example, we restrict access to the de-identified data using encryption and access control mechanisms, and require scientists and engineers to go through appropriate privacy and healthcare regulation related trainings before being granted access to the de-identified data.

Given the sensitivity of De-ID data pipelines, we do not release the prompt verbatim or source code of \ourmodel\ to prevent potential privacy risks and attacks. However, to support transparency and reproducibility, we provide descriptions of individual components, including retrieval-based relexicalization, hybrid rule/LLM logic, intelligent routing, and a two-step audio redaction process,  in the pipeline. This approach enables secure replication of our methodology while safeguarding patient confidentiality in our system.

\subsection{Intended Use of De-Identified Data.}\label{sec:intended_use_deid_data}
The de-identified data produced by \ourmodel\ is intended for a variety of critical use cases within AI-driven healthcare systems. First, it serves as a valuable resource for \textit{debugging production issues}, enabling engineers and data scientists to analyze system behavior and identify root causes of errors without compromising patient privacy. Second, the data supports \textit{understanding production model behavior}, providing insights into model performance, biases, and failure modes, which guide iterative improvements and model refinements. Finally, the de-identified dataset can be leveraged for \textit{training and evaluating downstream machine learning models}, including clinical documentation automation, clinical named entity recognition, and ``needle in a haystack'' tasks such as rare condition detection or retrieval of highly specific information from longitudinal records. Additionally, \textit{de-identified data is essential for conducting R\&D and facilitates collaboration with external researchers and clinicians}, enabling innovation while ensuring compliance with privacy regulations. These applications illustrate the dual importance of ensuring privacy while maintaining data utility for real-world healthcare advancements.



\subsection{All Models}
\label{sec:app_other_models}
We compare our methods with others as follows:
\begin{enumerate}
    \item \citet{yashwanth2024zero}: This study uses a zero-shot approach with two prompts -- brief and detailed -- applied with GPT-3.5 and GPT-4. The model returns ``[Censored]'' in lieu of explicit entity labels. In our experiments, we evaluate this approach using the GPT-4o to be a fair comparison with the Auto De-ID model. 
    \item \citet{altalla2025evaluating}: This study employs a zero-shot prompt with GPT-3 and GPT-4, where outputs are marked ``[Redacted]'' rather than providing explicit entity annotations. Although originally evaluated on a proprietary dataset, we adapt this baseline for the i2b2 corpus and evaluate it using the GPT-4o variant.
    \item \ourmodel\ (ours): The Auto De-ID model’s outputs are post-processed to align with our predefined PHI/PII entity categories. We use the following configuration parameters to ensure consistency across experiments: max\_passes = 2, max\_words = 256, and temperature = 0.
    \item Commercial Cloud APIs:  We assess two widely adopted commercial De-ID services that offer API-based solutions for PHI/PII extraction, namely, AWS (Amazon Web Services Medical Comprehend) and JSL (John Snow Labs). For these services, we standardize the output entity tags to align with our evaluation schema, ensuring fair comparison across systems.
\end{enumerate}

These settings were chosen based on preliminary tuning to balance performance and computational efficiency.

\subsection{Auto De-ID Algorithm}
\label{sec:app_autodeid_algo}
\begin{algorithm}
\caption{Auto De-ID Algorithm}
\label{algo:auto_deid}
\begin{algorithmic}[1]
\Require Text $T$, chunk size $\omega$, prediction model $M$, entity types $\mathcal{E}$, passes $p$
\Ensure Redacted text $\hat{T}$
\State Split $T$ into $m = \lceil |T| / \omega \rceil$ chunks.
\For{each chunk $c_i$}
    \For{$j = 0$ to $p-1$}
        \State Extract entities $\mathcal{R}_i \gets M(c_i, \mathcal{E})$
        \State Update fact dictionary $\mathcal{D} \gets \mathcal{D} \cup \mathcal{R}_i$
    \EndFor
\EndFor
\State Replace detected entities in $T$ with placeholders.
\Return $\hat{T}$
\end{algorithmic}
\end{algorithm}

The Auto De-ID algorithm processes text $T$ by segmenting it into $m = \lceil |T| / \omega \rceil$ chunks. Each chunk undergoes $p$ passes of entity extraction, where the set of identified entities is aggregated:
\[ \mathcal{D} = \bigcup_{i=1}^{m} \bigcup_{j=1}^{p} \mathcal{R}_i \]

\textbf{Choice of Chunk Size:} The chunk size $\omega$ is chosen to balance entity extraction accuracy. Sending large texts in a single request increases the risk of missing entities, as the model may fail to attend to all parts equally. By processing smaller chunks, we reduce the chance of entity leakage and improve recall.

\textbf{Multiple Passes Strategy:} Using multiple passes ($p$) helps mitigate biases in the model’s attention. In the first pass, the model extracts easily detectable entities. In the subsequent passes, previously detected entities are masked using their entity type (e.g., \texttt{[PERSON]}), allowing the model to focus on overlooked entities. This iterative process enhances recall, especially for underrepresented entity types. These hyperparameters can be adjusted based on the capabilities of the chosen LLM.

\textbf{Handling Context in Redaction:}  
A naive approach to redaction might simply replace all extracted entity mentions in the text with their corresponding entity types (e.g., replacing ``76'' with \texttt{[AGE]}). However, this often leads to over-redaction when identical strings appear in different contexts. Consider the sentence:

\begin{quote}
\textit{``The patient is 76 years old and takes 76 mg of aspirin daily.''}
\end{quote}

In this case, only the first occurrence of ``76'' refers to the patient's age and should be redacted as \texttt{[AGE]}. The second occurrence of ``76'' is part of a medication dosage and should not be redacted as age. If we blindly replace all instances of ``76'' with \texttt{[AGE]}, we would incorrectly redact ``76 mg,'' resulting in a loss of valuable clinical information.

Our entity extraction method avoids this by instructing the LLM to extract entities along with sufficient context that signals their specific meaning and position in the text. In this example, ``76 years old'' would be extracted as an \texttt{[AGE]} entity, while ``76 mg'' would either be ignored or extracted as a separate \texttt{[DOSAGE]} entity. This ensures that only the appropriate mention is redacted.

Since obtaining exact character positions from LLMs is unreliable, context-based entity extraction allows us to align each detected entity with its precise occurrence in the text, ensuring accurate and minimal redaction.

Detected entities in $T$ are replaced with placeholders to yield the final redacted text $\hat{T}$, ensuring sensitive data is masked correctly while preserving non-sensitive content.

\subsubsection{Example Workflow}
\label{sec:app_auto_deid_example}

To illustrate Auto De-ID’s process, consider the following structured JSON input:

\noindent\begin{small}
\begin{tabularx}{\columnwidth}{|l|X|}
\hline
\textbf{Field} & \textbf{Value} \\
\hline
\texttt{patient\_name} & Robert Johnson \\
\texttt{patient\_id} & A12345 \\
\texttt{gender} & male \\
\texttt{medical\_history} & Robert Johnson, a patient aged 53 was admitted to Springfield General Hospital for chest pain. Dr. Mary Smith prescribed medication. \\
\hline
\end{tabularx}
\end{small}

\textbf{Step 0: Schema Identification} The De-ID schema specifies rules for each field:
\vspace{2mm}

\noindent\begin{small}
\begin{tabularx}{\columnwidth}{|l|X|}
\hline
\textbf{Field} & \textbf{De-ID Rule} \\
\hline
\texttt{patient\_name} & Mask as \texttt{[PERSON]} \\
\texttt{patient\_id} & Hash \\
\texttt{gender} & Pass-through \\
\texttt{medical\_history} & Auto De-ID (LLM-based) \\
\hline
\end{tabularx}
\end{small}

\textbf{Step 1: Schema Processing} The structured fields are processed:
\vspace{2mm}

\noindent\begin{small}
\begin{tabularx}{\columnwidth}{|l|X|}
\hline
\textbf{Field} & \textbf{Processed Value} \\
\hline
\texttt{patient\_name} & \texttt{[PERSON]} \\
\texttt{patient\_id} & \texttt{HASH(A12345)} \\
\texttt{gender} & male \\
\hline
\end{tabularx}
\end{small}

\textbf{Step 2: Chunking} The unstructured text is split into overlapping chunks:
\vspace{2mm}

\noindent\begin{small}
\begin{tabularx}{\columnwidth}{|l|X|}
\hline
\textbf{Chunk} & \textbf{Text} \\
\hline
C1 & "Robert Johnson, a patient aged 53 was admitted to Springfield General Hospital for chest pain." \\
C2 & "for chest pain. Dr. Mary Smith prescribed medication." \\
\hline
\end{tabularx}
\end{small}

\textbf{Step 3: LLM-Based Entity Extraction} The model extracts PHI:
\vspace{2mm}

\noindent\begin{small}
\begin{tabularx}{\columnwidth}{|l|X|}
\hline
\textbf{Entity Type} & \textbf{Extracted Entities} \\
\hline
PERSON & Robert Johnson, Dr. Mary Smith \\
ORGANIZATION & Springfield General Hospital \\
\hline
\end{tabularx}
\end{small}

\textbf{Step 4: Multi-Pass Refinement} Previously detected entities are masked in subsequent passes:
\vspace{2mm}

\noindent\begin{small}
\begin{tabularx}{\columnwidth}{|l|X|}
\hline
\textbf{Pass} & \textbf{Input Chunk for 2nd Pass After Masking} \\
\hline
2nd & "[PERSON], a patient aged 53 was admitted to [ORGANIZATION] for chest pain." \\
2nd & "for chest pain. [PERSON] prescribed medication." \\
\hline
\end{tabularx}
\end{small}

\textbf{Step 5: Final Entity Extraction} Previously detected entities combined with the second pass extractions.
\vspace{2mm}

\noindent\begin{small}
\begin{tabularx}{\columnwidth}{|l|X|}
\hline
\textbf{Entity Type} & \textbf{Extracted Entities} \\
\hline
PERSON & Robert Johnson, Dr. Mary Smith \\
ORGANIZATION & Springfield General Hospital \\
AGE & 53 \\
\hline
\end{tabularx}
\end{small}

\textbf{Step 5: Final Redaction.} The final de-identified record:
\vspace{2mm}

\noindent\begin{small}
\begin{tabularx}{\columnwidth}{|l|X|}
\hline
\textbf{Field} & \textbf{Final Value} \\
\hline
\texttt{patient\_name} & \texttt{[PERSON]} \\
\texttt{patient\_id} & \texttt{HASH(A12345)} \\
\texttt{gender} & male \\
\texttt{medical\_history} & "[PERSON], a patient aged [AGE] was admitted to [ORGANIZATION] for chest pain. [PERSON] prescribed medication." \\
\hline
\end{tabularx}
\end{small}

This final output ensures all PHI/PII is masked while maintaining text coherence.

\subsection{Audio De-ID Algorithm}
\label{sec:app_audio_deid}
\begin{figure}[htbp]
    \centering
    \includegraphics[width=\columnwidth]{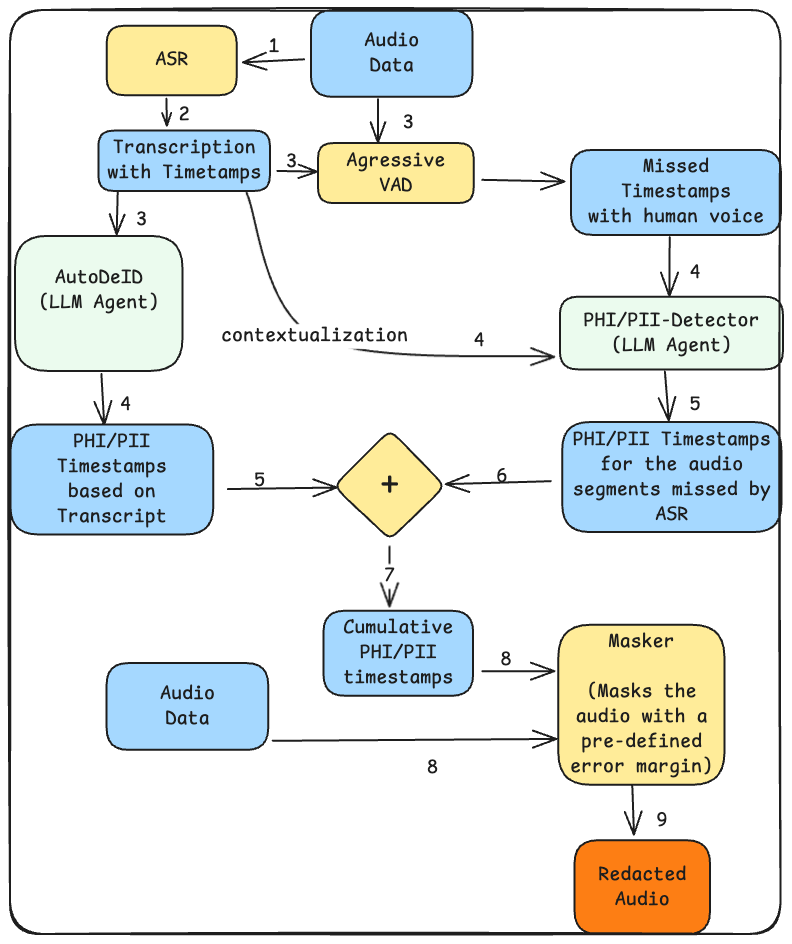}
    \caption{Audio De-Id Workflow Diagram}
    \label{fig:audiodeid_workflow}
\end{figure}

\begin{algorithm}
\caption{Audio De-ID Algorithm}
\label{alg:audiod}
\begin{algorithmic}[1]
\Require Audio $A$, ASR model $M_{ASR}$, VAD $M_{VAD}$, De-ID model $M_{Deid}$, PHI/PII Detector $M_{LLM}$, entity types $\mathcal{E}$
\Ensure Redacted Audio $A_{\hat{T}}$
\State Generate transcript $T \gets M_{ASR}(A)$
\State Extract PHI/PII $\mathcal{D} \gets M_{Deid}(T, \mathcal{E})$
\State Identify missing timestamps $T_{\text{missing}}$ and detect human speech with $M_{VAD}$
\For{each human-voiced timestamp $t_{\text{human}}$}
    \State Extract context, detect PHI/PII with $M_{LLM}$, and update $\mathcal{D}$
\EndFor
\State Mute detected PHI/PII in $A$.
\Return $A_{\hat{T}}$
\end{algorithmic}
\end{algorithm}

Audio De-ID first converts speech to text using ASR:
\[ T = M_{ASR}(A) \]
Entities are extracted from $T$ to construct $\mathcal{D}$:
\[ \mathcal{D} = M_{Deid}(T, \mathcal{E}) \]
Gaps in ASR timestamps $T_{\text{missing}}$ are analyzed with $M_{VAD}$ to identify human speech regions, where PHI/PII detection is refined using an LLM-based model. The final redacted audio $A_{\hat{T}}$ is generated by muting PHI-containing segments.

\subsubsection{Example Workflow}
\label{sec:app_audio_deid_example}

\textbf{ASR-Generated Transcript:}
\noindent\texttt{
\begin{quote}
``The patient visited Dr. Smith last week a follow-up in his clinic at Creekwood Hospital. They discussed medication changes and scheduled the next appointment for next month. The patient also mentioned feeling unwell over the weekend.''
\end{quote}
}
\textbf{Auto De-ID Detected PHI:}
\begin{itemize}
    \item ``Dr. Smith'' (00:04.23 - 00:04.80)
    \item ``Creekwood Hospital'' (00:12.57 - 00:13.20)
\end{itemize}

\textbf{Identifying Missing Timestamps (Set Subtraction):}
\begin{itemize}
    \item (00:02.85 - 00:02.95) (Missed speech)
    \item (00:07.42 - 00:07.57) (Missed speech)
    \item (00:15.10 - 00:15.30) (Missed speech)
\end{itemize}

\textbf{VAD Filtering (Keeping Only Human Speech Segments):}
\begin{itemize}
    \item (00:02.85 - 00:02.95) - Human voice detected $\checkmark$
    \item (00:07.42 - 00:07.57) - Human voice detected $\checkmark$
    \item (00:15.10 - 00:15.30) - Background noise, discarded $\times$
\end{itemize}

\textbf{Timestamp Adjustment for ASR Errors:}  
To compensate for ASR errors, timestamps are adjusted with a safe margin of 300ms:
\begin{itemize}
    \item \textbf{Before:} ``Dr. Smith'' (00:04.23 - 00:04.80)
    \item \textbf{After:} ``Dr. Smith'' (00:03.93 - 00:05.10)
    \item \textbf{Before:} ``Creekwood Hospital'' (00:12.57 - 00:13.20)
    \item \textbf{After:} ``Creekwood Hospital'' (00:12.27 - 00:13.50)
\end{itemize}

\textbf{Reconstructing the Transcript:}  
The transcript is updated by inserting missing timestamps:

\noindent\texttt{
\begin{quote}
``\textbf{$<$human\_timestamp\_(00:02.85 - 00:02.95)$>$} The patient visited Dr. Smith last week \textbf{$<$human\_timestamp\_(00:07.42 - 00:07.57)$>$} a follow-up in his clinic at Creekwood Hospital. They discussed medication changes and scheduled the next appointment for next month. The patient also mentioned feeling unwell over the weekend.''
\end{quote}
}
This updated transcript is analyzed by an LLM to predict the likelihood of inserted timestamps containing PHI/PII. 
For eg., Lets say the LLM predicted the following:
\begin{itemize}
    \item \textbf{$<$human\_timestamp\_(00:02.85 - 00:02.95)$>$} - NON-PHI/PII
    \item \textbf{$<$human\_timestamp\_(00:07.42 - 00:07.57)$>$} - PHI/PII
\end{itemize}
Then the final set of detected PHI/PII timestamps—extracted via Auto De-ID and combined by the LLM—guided additional timestamps is used for muting the corresponding sections in the final redacted audio. The final audio will look like as follows:

\noindent\texttt{
\begin{quote}
``The patient visited [MUTED] last week [MUTED] a follow-up in his clinic at [MUTED]. They discussed medication changes and scheduled the next appointment for next month. The patient also mentioned feeling unwell over the weekend.''
\end{quote}
}

\subsection{Auto Relexicalization Algorithm}
\label{sec:app_auto_relex}

\begin{figure*}[t]
    \centering
    \includegraphics[width=\textwidth]{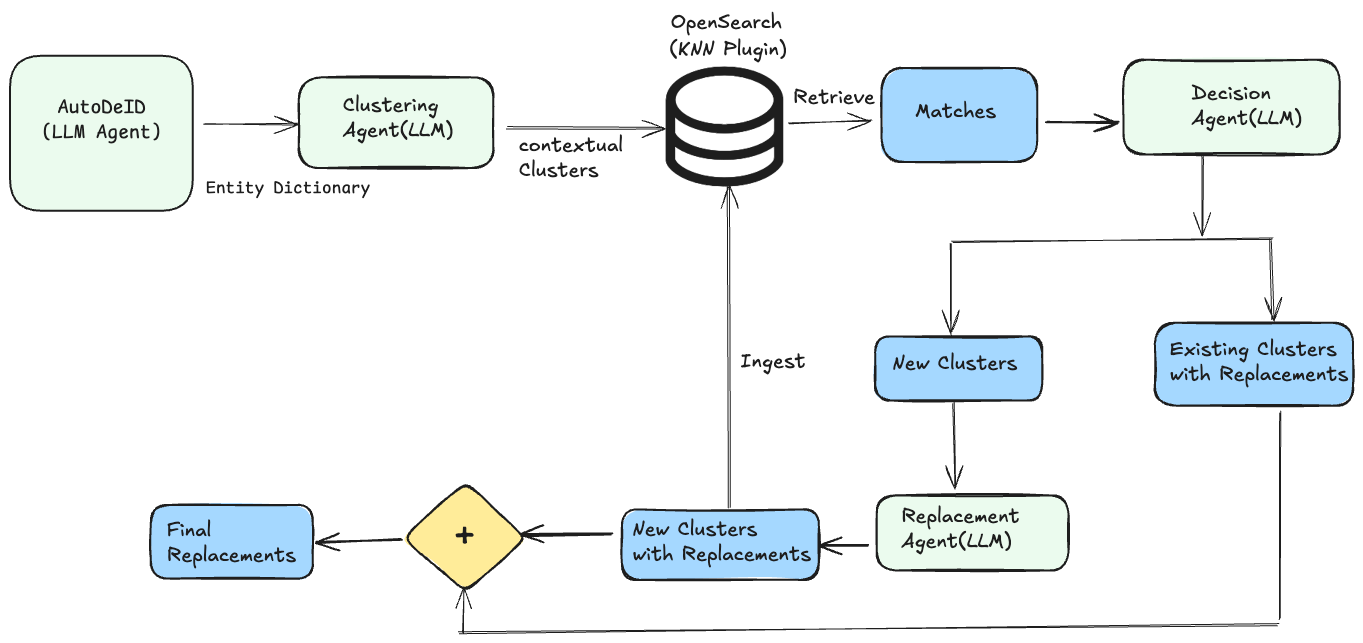}
    \caption{Relexicalization Workflow Diagram}
    \label{fig:relex_workflow}
\end{figure*}

\begin{algorithm}
\caption{Auto Relexicalization Algorithm}
\label{alg:relexicalization}
\begin{algorithmic}[1]
\Require Text $T$, fact dictionary $\mathcal{D}$, index $I$, clustering $M_{\text{cluster}}$, retrieval $M_{\text{search}}$, decision $M_{\text{decision}}$, replacement model $M_{\text{replace}}$
\Ensure Relexicalized text $\hat{T}$
\State Cluster entities: $C_{\mathcal{D}} \gets M_{\text{cluster}}(T, \mathcal{D})$
\For{each cluster $C_i$}
    \State Generate query $Q_i$ and retrieve match $R_i \gets M_{\text{search}}(Q_i, I)$
    \If{Valid replacement $M_{\text{decision}}(Q_i, R_i, T)$}
        \State Use $R_i$
    \Else
        \State Generate new replacement $R_{\text{new}} \gets M_{\text{replace}}(Q_i, T)$
        \State Ingest new replacement and original entity into index $I$
        \State Store $R_{\text{new}}$ for final relexicalization
    \EndIf
\EndFor
\State Apply adjustments and replace entities.
\Return $\hat{T}$
\end{algorithmic}
\end{algorithm}

Auto Relexicalization clusters fact entities using $M_{\text{cluster}}$ and retrieves candidate replacements from an index using vector search:
\[ R_i = M_{\text{search}}(Q_i, I) \]
If the decision model $M_{\text{decision}}$ rejects the match, a new replacement $R_{\text{new}}$ is generated using a replacement model:
\[ R_{\text{new}} = M_{\text{replace}}(Q_i, T) \]
This new replacement, along with the original entity, is then ingested into the index $I$ to ensure consistency across documents. The final text $\hat{T}$ is formed after replacing entities accordingly.

\subsection{Proposed Metrics for Relexicalization}
\label{tab:relex_metrics}
We propose a set of evaluation metrics to assess the effectiveness of re-lexicalization in preserving entity roles, maintaining contextual coherence, ensuring replacement consistency, and minimizing unintended biases in clinical models.

\noindent\textbf{Entity Preservation Rate} evaluates whether the re-lexicalized entity retains its semantic role and contextual attributes. Higher scores indicate better preservation. For instance, in the sentence \textit{“Dr. Emily Carter is a cardiologist at St. Mary’s Hospital,”} a poor re-lexicalization would be \textit{“Alice is a teacher at Westwood Academy”}, which alters both the profession and institution type. A good re-lexicalization would be \textit{“Dr. Kevin Chang is a cardiologist at Lincoln Medical Center”}, as it preserves the entity’s role and contextual relevance.\\

\noindent\textbf{Contextual Coherence Score} measures whether the re-lexicalized entity integrates naturally within the surrounding text without disrupting fluency or meaning. For example, in the original sentence \textit{“John met his lawyer, Mr. Anderson, at the firm,”} a poor substitution would be \textit{“Harry met his lawyer, Pizza Hut, at the firm”}, introducing a semantic inconsistency. A more appropriate replacement would be \textit{“Harry met his lawyer, Mr. Bennett, at the firm”}, maintaining contextual coherence.\\

\noindent\textbf{Replacement Consistency Score} ensures that an entity is consistently replaced across multiple documents, preserving identity coherence. For instance, in \textit{“Dr. Emily Carter attended the surgery”}, a conflicting replacement in another document such as \textit{“Dr. Jennifer Smith is a cardiologist”} introduces inconsistency. A high score indicates that the same entity is replaced uniformly across contexts.\\

\noindent\textbf{Clinical Model Consistency} assesses whether re-lexicalized data, when used in clinical decision-making models, avoids introducing biases related to race, ethnicity, region, or age group. If a model trained on real data produces a metric value \( X \), a poor re-lexicalization may yield a metric of \( X + \delta{x} \), where \( \delta{x} \) is significantly large, indicating a deviation from real-world behavior. An optimal re-lexicalization ensures that the metric shift remains marginal, preserving the integrity of the clinical model.

\begin{figure*}[t]
    \centering
    \includegraphics[width=\textwidth]{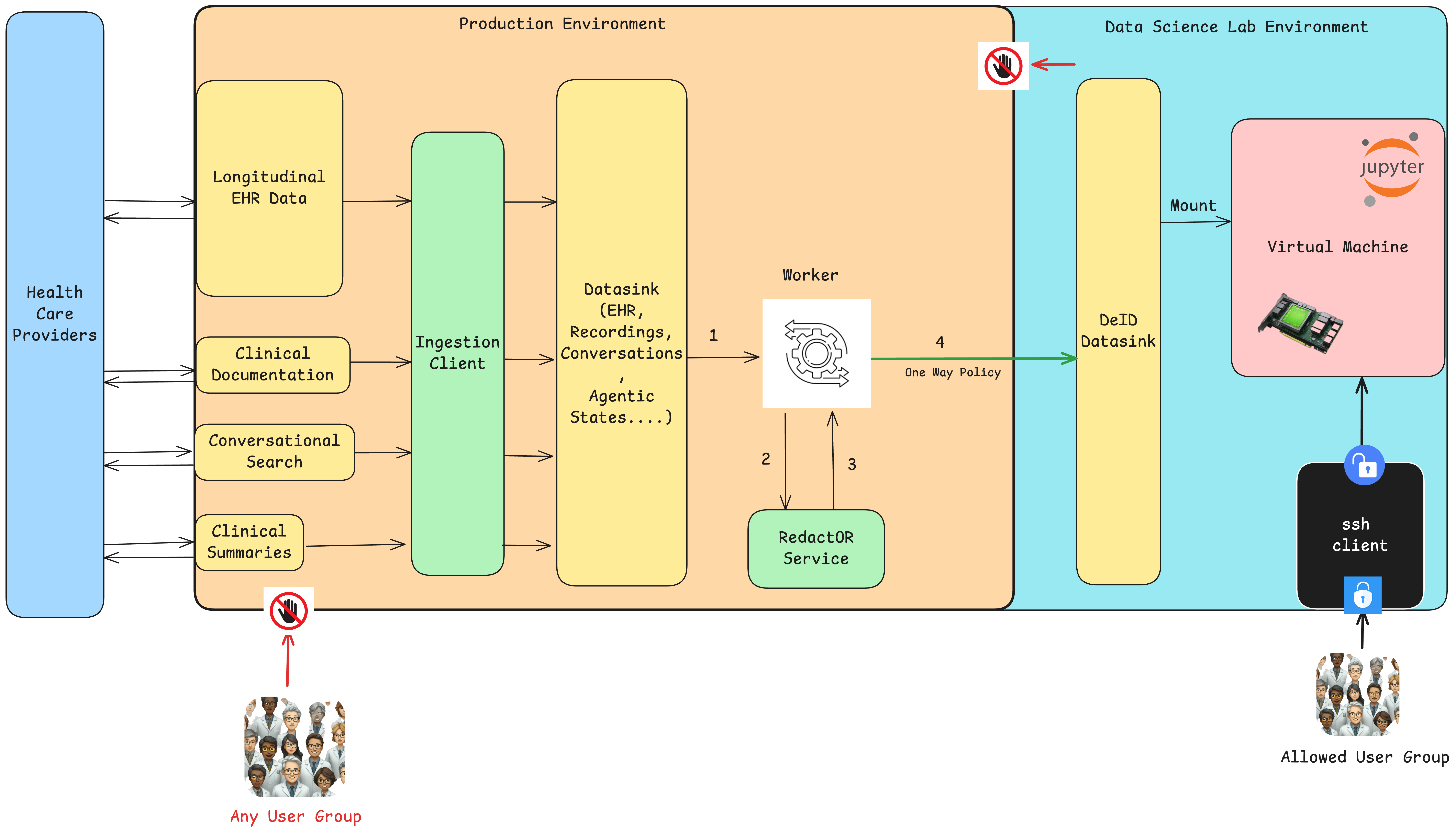}
    \caption{Diagram explaining the integration of our \ourmodel\ framework with Oracle Health Clinical AI System}
    \label{fig:integration_workflow}
\end{figure*}

\newpage

\subsection{Integration of the \ourmodel\ with Oracle Health \& Clinical AI System}
\label{sec:system_integration}

Our \ourmodel\ framework is integrated into Oracle Health Clinical AI system to facilitate the privacy-preserving processing of longitudinal EHRs, ambient intelligence data, and conversational AI outputs. The system operates autonomously, as the Production Environment is inaccessible to any user group, ensuring a fully automated pipeline. A dedicated worker module, running continuously, monitors the Datasink, retrieves unprocessed files, submits them to the \ourmodel\ service for PHI/PII removal, and securely transfers the de-identified records to the Data Science Lab Environment. A one-way policy enforces strict data flow control, guaranteeing that only de-identified data is accessible within a secure research environment, where authorized users interact with it via an SSH-secured virtual machine, preserving both data integrity and analytical utility.

\clearpage
\onecolumn

\subsection{End to End Example Illustrating Text De-identification and Relexicalization using \ourmodel}
\label{tab:e2e_example}
\subsubsection{Schema for the Input Record}
\label{schema_definition_example}
\begin{lstlisting}[caption={Sample Schema Definition}]
schema_definition = {
  "$schema": "http://json-schema.org/draft-04/schema#",
  "type": "object",
  "recordVersion": "1.0",
  "description": "Schema for a free text record",
  "dataType": "clinicalRecord"
  "properties": {
    "PatientId": {
      "type": "string",
      "description": "Patient ID.",
      "autoDeId": False,
      "shouldMask": False,
      "shouldHash": True,
      "entity_type": None
    },

    "MRN": {
      "type": "string",
      "description": "MRN of the Patient",
      "autoDeId": False,
      "shouldMask": False,
      "shouldHash": True,
      "entity_type": None
    },

    "AGE": {
      "type": "string",
      "description": "Age of the Patient",
      "autoDeId": False,
      "shouldMask": True,
      "shouldHash": False,
      "entity_type": "[AGE]"
    },

    "note": {
      "type": "string",
      "description": "Clinical Note",
      "autoDeId": True,
      "shouldMask": False,
      "shouldHash": False,
      "entity_type": None
    }
  },
}
\end{lstlisting}

\subsubsection{Input Record}
\begin{lstlisting}[caption={Sample record with PHI/PII}]
{
  "PatientId": "123456789",
  "MRN": "A987654321",
  "AGE": "45 years",
  "note": "John Doe, a 45-year-old male, presented to Stanford Medical Center on 03/16/2025 complaining of severe abdominal pain. He was referred by Dr. Emily Smith from Valley Health Clinic. His address is 123 Main St, Palo Alto, CA 94301. Contact number: (650) 555-1234. Past medical history includes hypertension and Type 2 diabetes. His insurance ID is INS-789456123. The patient's wife, Jane Doe, can be reached at (650) 555-5678. A CT scan was performed and results were discussed with the patient at 2:00 PM. Follow-up scheduled on 03/22/2025 at 9:00 AM. Patient is employed as a software engineer at TechNova Corp. Social Security Number: 987-65-4321."
}
\end{lstlisting}

\subsubsection{Entities Extracted by Auto De-ID}
\begin{lstlisting}[caption={Extracted Entities}]
{
    "PERSON": [
        "Jane Doe",
        "Emily Smith",
        "John Doe"
    ],
    "ADDRESS": [
        "123 Main St, Palo Alto, CA 94301"
    ],
    "AGE": [
        "45 years"
        "45-year-old"
    ],
    "LOCATION": [
        "Palo Alto"
    ],
    "MARITAL_STATUS": [
        "wife"
    ],
    "PARENTHOOD": [],
    "OCCUPATION": [
        "software engineer"
    ],
    "BIRTH_DATE_TIME": [],
    "SSN_OR_TAXPAYER": [
        "987-65-4321"
    ],
    "EMAIL": [],
    "FIN": [
        "INS-789456123"
    ],
    "GUID": [
        "987-65-4321"
    ],
    "ORGANIZATION": [
        "TechNova Corp",
        "Stanford Medical Center",
        "Valley Health Clinic"
    ],
    "PHARMACY": [],
    "DIAGNOSTIC_LABS": []
}
\end{lstlisting}
\newpage

\subsubsection{DeID-Only Output}
\begin{lstlisting}[caption={DeID-Only Output}]
{
  "PatientId": "HASHED_VALUE",
  "MRN": "HASHED_VALUE",
  "AGE": "[AGE]",
  "note": "[PERSON], a [AGE] male, presented to [ORGANIZATION] on 03/16/2025 complaining of severe abdominal pain. He was referred by Dr. [PERSON] from [ORGANIZATION]. His address is [ADDRESS]. Contact number: [TELEPHONE_NUMBER]. Past medical history includes hypertension and Type 2 diabetes. His insurance ID is [FIN]. The patient's [MARITAL_STATUS], [PERSON], can be reached at [TELEPHONE_NUMBER]. A CT scan was performed and results were discussed with the patient at 2:00 PM. Follow-up scheduled on 03/22/2025 at 9:00 AM. Patient is employed as a [OCCUPATION] at [ORGANIZATION]. Social Security Number: [GUID]."
}
\end{lstlisting}

\subsubsection{De-ID + Relexicalization Output}
\begin{lstlisting}[caption={DeID+Relexicalization Output}]
{
  "PatientId": "HASHED_VALUE",
  "MRN": "HASHED_VALUE",
  "AGE": "mid-forties",
  "note": "Michael Johnson, a mid-forties male, presented to Harvard Medical Center on 03/16/2025 complaining of severe abdominal pain. He was referred by Dr. Sophia Brown from Green Valley Clinic. His address is 456 Elm St, Mountain View, CA 94041. Contact number: (123) 274-0846. Past medical history includes hypertension and Type 2 diabetes. His insurance ID is INS-123456789. The patient's spouse, Alice Johnson, can be reached at (123) 274-6354. A CT scan was performed and results were discussed with the patient at 2:00 PM. Follow-up scheduled on 03/22/2025 at 9:00 AM. Patient is employed as a data scientist at Innovatech Inc.. Social Security Number: 123-45-6789."
}
\end{lstlisting}

\newpage
\section{High Level Prompt Templates for Different LLM Components}

Due to compliance, privacy, and business confidentiality considerations, we do not release the exact prompts used for each LLM component in \ourmodel. Instead, to foster reproducibility and enable community adaptation, we provide high-level prompt templates that capture the \textit{structure, intent, and output format} of each prompt while omitting sensitive implementation details.

These templates define:
\begin{itemize}
    \item The \textbf{role of the LLM} in each component (e.g., entity extraction, clustering, relexicalization),
    \item The \textbf{core task description} and \textbf{guidelines} for execution,
    \item The \textbf{expected output schema} in JSON format for integration and evaluation,
    \item \textbf{Placeholders} for data inputs, reference context, and special parameters (e.g., shift values, entity-specific rules).
\end{itemize}

By offering these templates, we enable researchers and practitioners to develop specialized prompts tailored to their own datasets, privacy policies, and LLM configurations, while ensuring compatibility with our overall system architecture.

\subsection{Auto De-ID LLM Component}
\label{tab:deid_prompt_templates}
\noindent\begin{minipage}{\linewidth}
\begin{Verbatim}
prompt: |
  {{role}}  # High-level role of the LLM. 
  Describe that it acts as a De-Identification Specialist tasked with 
  extracting PHI/PII from clinical text while adhering to legal privacy regulations 
  (e.g., HIPAA). Include general expectations on accuracy and coverage.

  {{entity_extraction_guidelines}}  # Instructions on how to treat text 
  (e.g., case sensitivity, contractions), exclusions (e.g., medication names, diagnoses),
  and how to handle special cases. Mention precision requirements for each entity type 
  and the need to distinguish similar types (e.g., ADDRESS vs LOCATION).

  {{controls}}  # List and description of specific PHI/PII categories to be extracted 
  (e.g., names, dates, contact info, IDs, financial details, technical identifiers, 
  demographic information). This section should align with the entity types and provide 
  guidance on inclusion/exclusion criteria.

  {{context_awareness}}  # Explain the need for context-sensitive entity extraction to 
  avoid over-redaction. Describe how identical strings may appear in different contexts 
  with different meanings and the importance of using surrounding context to correctly 
  identify which instance to redact. Highlight that the model must associate each extracted 
  entity with its specific textual occurrence based on context, not just string matching.

  The output format should strictly just be a JSON dictionary with the entity mentioned 
  above as the key and its list of words/phrases found in the text as its value.

  For eg., "ENTITY": ["A", "B", "C"]

  You must not add any key which is not a part of the guidelines above. You must add all 
  the entity as the keys in the output even if the value list for that is empty.

  The final output format must look like as follows. You must not produce anything 
  except the json output. Ensure the output can be parsed by Python json.loads

  {
    <entity_type1>: <list_of_words_or_phrases_for_entity_type1>,
    <entity_type2>: <list_of_words_or_phrases_for_entity_type2>,
    ... and so on
  }

  {{self_checklist}}  # List of validation checks the LLM must perform before returning output. 
  For example, ensure PERSON doesn’t include pronouns, validate that BIRTH_DATE_TIME only includes
  birth dates, and confirm all keys in output match allowed entity types.

  Here is the input text:
  {{input_text}}  # Placeholder for the input clinical text to be de-identified. 
  This is the text the LLM will process.
\end{Verbatim}
\end{minipage}

\newpage
\subsection{Relexicalizer Components}
\label{tab:relex_prompt_templates}
\noindent\begin{minipage}{\linewidth}
\begin{Verbatim}
clustering_prompt: |
  {{role}}  # Describe the clustering task: grouping contextually identical entities from two medical documents.

  {{task_overview}}  # Outline the goal: assign consistent identifiers to contextually similar entities across documents.

  {{guidelines}}  # Provide detailed guidelines: consistency, identifier format, handling of subnames, JSON format compliance.

  {{example_input_output}}  # Include sample input/output structure for clarity (optional for template use).

  {{input_dict_placeholder}}  # Placeholder for the input dictionary of entities.

  {{reference_text_placeholder}}  # Placeholder for the contextual text reference.

  Output Format:
  ```json
  {
    "ENTITY_TYPE": {
      "ENTITY_TYPE_1": ["entity_variant_1", "entity_variant_2"],
      "ENTITY_TYPE_2": ["entity_variant_3"]
    },
    ...
  }
  
query_prompt: |
  {{role}}  # Describe task: generate semantic query strings for each entity cluster to retrieve similar entities.

  {{guidelines}}  # Explain how to synthesize cluster information into a concise semantic query.

  {{example_input_output}}  # Provide example input and expected query outputs for context (optional for template use).

  {{cluster_placeholder}}  # Placeholder for the input clusters.

  {{reference_text_placeholder}}  # Placeholder for reference context.

  Output Format:
  ```json
  {
    "ENTITY_TYPE": {
      "ENTITY_TYPE_1": "query_string_for_entity_1",
      "ENTITY_TYPE_2": "query_string_for_entity_2"
    },
    ...
  }

decision_prompt: |
  {{role}}  # Describe task: evaluate semantic similarity of search results to query entities.

  {{guidelines}}  # Provide detailed decision rules for matching: exact match, partial, cultural context, ambiguity.

  {{constraints}}  # State output constraints: JSON format, result length consistency, no assumptions.

  {{example_input_output}}  # Include examples of query, search results, context, and expected Y/N output.

  {{query_placeholder}}  # Placeholder for input query.

  {{search_result_placeholder}}  # Placeholder for search results.

  {{context_placeholder}}  # Placeholder for reference context.

  Output Format:
  ```json
  {
    "result": ["Y", "N", ...]
  }

replacement_prompt: |
  {{role}}  # Describe task: generate realistic replacement values for each cluster entity.

  {{guidelines}}  # Explain general rules for replacement: alignment with entity attributes, format, and uniqueness.

  {{special_rules}}  # Detailed per-entity replacement rules for privacy-preserving, contextually realistic generation.

  Output Format:
  ```json
  {
    "ENTITY_TYPE_1": {"replacement": "replacement_value_1", "type": "ENTITY_TYPE"},
    "ENTITY_TYPE_2": {"replacement": "replacement_value_2", "type": "ENTITY_TYPE"},
    ...
  }
  ```
  {{input_cluster_placeholder}}  # Placeholder for the entity clusters to replace.
  {{existing_replacements_placeholder}}  # Placeholder for existing replacements to avoid duplication.
  {{reference_text_placeholder}}  # Placeholder for context to guide realistic replacements.

\end{Verbatim}
\end{minipage}

\subsection{Audio PHI/PII Timestamps Detector}
\label{tab:audio_prompt_templates}
\noindent\begin{minipage}{\linewidth}
\begin{Verbatim}
prompt: |
  {{role}}  # Describe task: infer most likely entity type 
  for missing audio segments using surrounding transcript context.

  {{entity_definitions}}  # Provide detailed descriptions of each possible 
  entity type (e.g., PERSON, AGE, ADDRESS, etc.) and how to identify them from context.

  {{transcript_format}}  # Describe how transcript data is presented
  with timestamps and missing sections.

  {{example_input_output}}  # Provide an example transcript with missing
  timestamps and corresponding expected entity predictions.

  {{task_instruction}}  # Instruct model to return a JSON dictionary where
  keys are timestamps and values are predicted entity types or "UNKNOWN".

  Output Format:
  ```json
  {
    "TIMESTAMP1": "ENTITY_TYPE",
    "TIMESTAMP2": "ENTITY_TYPE",
    ...
  }
  ```

  {{partial_transcript_placeholder}}  # Placeholder for the transcript 
  input with missing sections.
\end{Verbatim}
\end{minipage}

\section{Entity types supported by RedactOR in Production}
\label{tab:entities}
\noindent\begin{minipage}{\linewidth}
\begin{Verbatim}
- ADDRESS
- SSN_OR_TAXPAYER
- EMAIL
- PASSPORT_NUMBER_US
- TELEPHONE_NUMBER
- DRIVER_ID_US
- BANK_ACCOUNT_NUMBER
- BANK_SWIFT
- BANK_ROUTING
- CREDIT_DEBIT_NUMBER
- MEDICAL_RECORD_NUMBER
- HEALTH_PLAN_ID
- CERTIFICATE_NUMBER
- FIN
- VEHICLE_LICENSE_PLATE_US
- VEHICLE_IDENTIFIER_US
- GUID
- PERSON
- DIAGNOSTIC_LABS
- PHARMACY
- ORGANIZATION
- AGE
- LOCATION
- PARENTHOOD
- MARITAL_STATUS
- OCCUPATION
- RACE
- ETHNICITY
- BIRTH_DATE_TIME
- DEATH_DATE_TIME
- IP_ADDRESS
- URL
- MAC_ADDRESS
\end{Verbatim}
\end{minipage}




\end{document}